\pdfoutput=1

\documentclass[11pt]{article}

\usepackage{acl}

\usepackage{times}
\usepackage{latexsym}

\usepackage[T1]{fontenc}

\usepackage[utf8]{inputenc}

\usepackage{microtype}

\usepackage{tabularx}
\usepackage{color}
\usepackage{booktabs}
\usepackage{graphicx}
\usepackage{graphics}
\usepackage{adjustbox}

\newcommand{\xlmr}{\mbox{XLM-R}}
\newcommand{\xmod}{\mbox{X-MOD}}
\newcommand{\swissbert}{SwissBERT}
\newcommand{\swissner}{SwissNER}
\newcommand{\bertscore}{BERTScore}
\newcommand{\wikineural}{WikiNEuRal}
\newcommand{\xstance}{\mbox{\textsc{x}-stance}}

\title{\swissbert{}: The Multilingual Language Model for Switzerland}

 \author{
    {\bf Jannis Vamvas$^1$ \hspace{2mm}}
    {\bf Johannes Graën$^2$ \hspace{2mm}}
    {\bf Rico Sennrich$^1$} \\
    $^1$Department of Computational Linguistics, University of Zurich \\
    $^2$Linguistic Research Infrastructure, University of Zurich \\
    \texttt{johannes.graen@linguistik.uzh.ch}, \\
    \texttt{\{vamvas,sennrich\}@cl.uzh.ch}
 }

\begin{document}
\maketitle
\begin{abstract}
We present \swissbert{}, a masked language model created specifically for processing Switzerland-related text.
\swissbert{} is a pre-trained model that we adapted to news articles written in the national languages of Switzerland – German, French, Italian, and Romansh.
We evaluate \swissbert{} on natural language understanding tasks related to Switzerland and find that it tends to outperform previous models on these tasks, especially when processing contemporary news and/or Romansh Grischun.
Since \swissbert{} uses language adapters, it may be extended to Swiss German dialects in future work.
The model and our open-source code are publicly released at \url{https://github.com/ZurichNLP/swissbert}.
\end{abstract}

\section{Introduction}
Self-supervised learning for natural language processing~(NLP) has inspired the release of numerous language models, like BERT~\cite{devlin-etal-2019-bert}.
However, NLP researchers in Switzerland, a country with four national languages, are confronted by a unique language situation.
Individual models for German, French or Italian~[\citealp{chan-etal-2020-germans,martin-etal-2020-camembert,polignano2019alberto} etc.] are difficult to combine for multilingual tasks, and massively multilingual models such as \xlmr{}~\cite{conneau-etal-2020-unsupervised} do not focus on the multilingualism that is particular to Switzerland.
The fourth national language, Romansh, is not represented in a neural language model so far.

In this paper, we describe \swissbert{}, a model trained on more than 21~million Swiss news articles with a total of 12~billion tokens.
By combining articles in Swiss Standard German, French, Italian, and Romansh Grischun, we aim to create multilingual representations by implicitly exploiting common entities and events in the news.

\begin{figure}
  \centering
  \includegraphics[width=\linewidth]{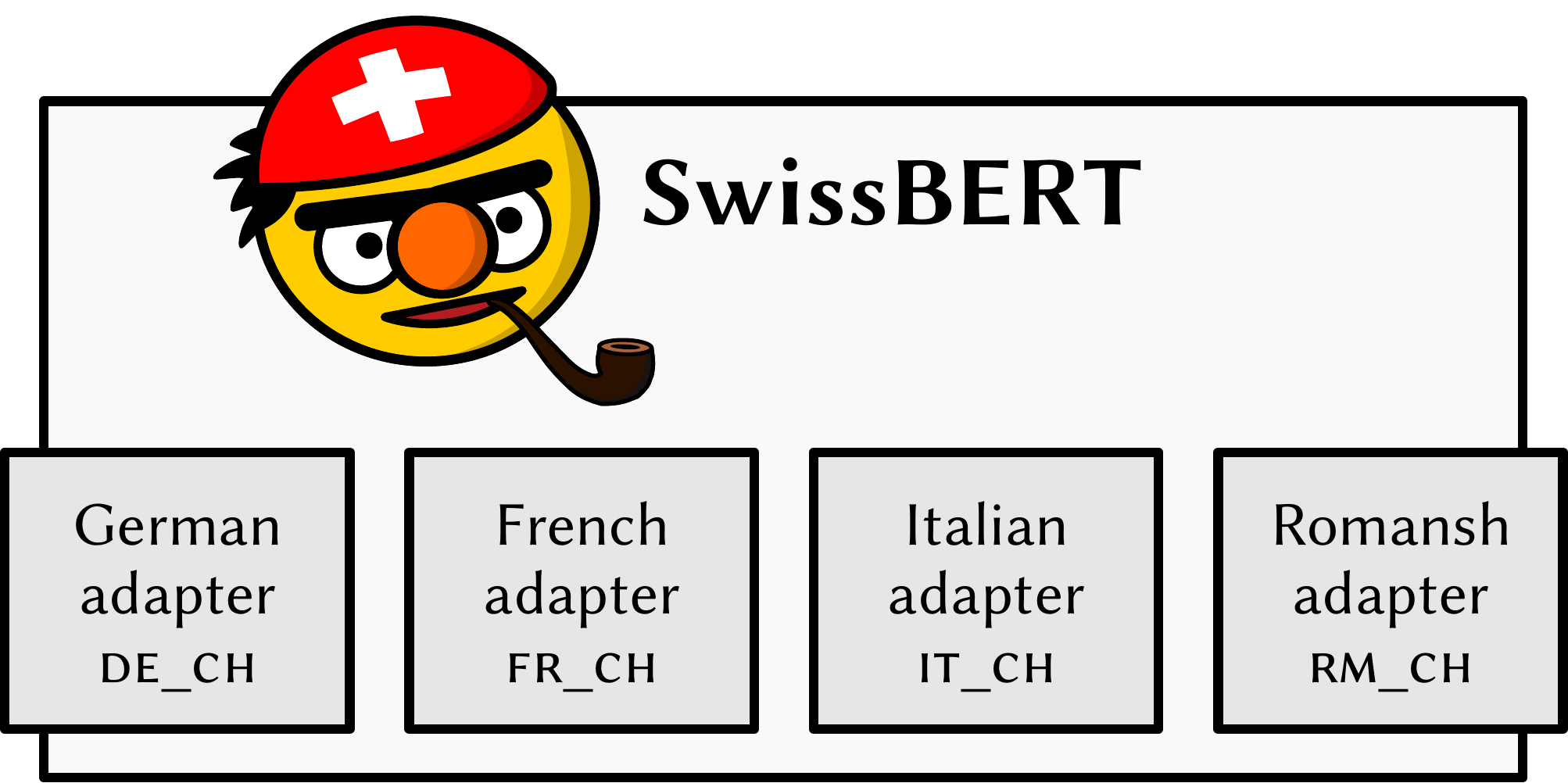}
  \caption{\swissbert{} is a transformer encoder with language adapters~\cite{pfeiffer-etal-2022-lifting} in each layer.
  There is an adapter for each national language of Switzerland.
  The other parameters in the model are shared among the four languages.}
\end{figure}

The \swissbert{} model is adapted from a \textit{Cross-lingual Modular} (\xmod{}) transformer that was pre-trained jointly in 81 languages~\cite{pfeiffer-etal-2022-lifting}.
We adapt \xmod{} to our corpus by training custom language adapters.
We also create a Switzerland-specific subword vocabulary for \swissbert{}.
The resulting model has 153M parameters.

Because \swissbert{} inherits \xmod{}'s modularity, future work may extend it beyond the four national languages.
In particular, Swiss German dialects are absent in our training corpus of written news articles but might have other resources that could be used for adding a fifth language adapter to \swissbert{}.

In order to evaluate our model, we create a test set for named entity recognition on contemporary news~(\mbox{\swissner{}}) and find that our model improves over common baselines.
When probing our model's capabilities on Romansh, we find that it strongly outperforms models that have not been trained on the language, both in terms of zero-shot cross-lingual transfer, and German--Romansh alignment of words and sentences~\cite{dolev-2023-romansh}.

Since \swissbert{} has been adapted to news articles only, we make sure to also gauge its out-of-domain performance.
We observe a moderate but systematic improvement over \xlmr{} when detecting stance in user-generated comments on Swiss politics~\cite{vamvas-sennrich-2020-xstance} but do not observe state-of-the-art accuracy when recognizing named entities in historical, OCR-processed news~\cite{ehrmann2022overview}.

We release the \swissbert{} model to the research community.\footnote{\url{https://huggingface.co/ZurichNLP/swissbert}}
Our code repository\footnote{\url{https://github.com/ZurichNLP/swissbert}} includes examples for fine-tuning on downstream tasks based on the \textit{transformers} library~\cite{wolf-etal-2020-transformers}.
Due to the nature of the pre-training corpus, the \swissbert{} model may currently not be used for commercial purposes.
However, our model may be used in any non-commercial setting, including academic research.

\section{Background and Related Work}

\paragraph{Masked Language Models}
Masked language modeling is a standard approach for learning computational representations from raw text.
Masked language models for various languages and domains have been released in the wake of the BERT model~\cite{devlin-etal-2019-bert}, a Transformer~\cite{vaswani2017transformer} that has been trained on English text.
For German, such monolingual models have been released by \citet{chan-etal-2020-germans} and \citet{DBLP:journals/corr/abs-2012-02110}, among others.
Similarly, monolingual masked language models have been created for French~\cite{martin-etal-2020-camembert,le-etal-2020-flaubert-unsupervised}, for Italian~\cite{polignano2019alberto,muffo2020bertino} and many other languages.
BERT-style models have also been trained on digitized historical newspapers~\cite{stefan_schweter_2020_4275044,schweter-et-al-2022-hmbert}.

\paragraph{Multilingual Models}
Some masked language models have been trained jointly on multiple languages, which allows for transfer learning across languages~\cite{devlin-etal-2019-bert,NEURIPS2019_c04c19c2}.
While massively multilingual models such as \xlmr{} enable transfer to languages that have fewer pre-training resources, their overall performance tends to decline compared to monolingual models~\cite{conneau-etal-2020-unsupervised}.
This trade-off extends to multilingual subword vocabularies that are created jointly for many languages and scripts~\cite{rust-etal-2021-good}.

\paragraph{Cross-lingual Modular Transformers}
\citet{pfeiffer-etal-2022-lifting} have proposed \xmod{}, a multilingual model that is similar to \xlmr{} but has monolingual components.
These components are included in each Transformer layer during pre-training.
In this paper, we refer to them as \textit{language adapters}, as they are reminiscent of adapters that are added post-hoc to a pre-trained model~\cite{pmlr-v97-houlsby19a,pfeiffer-etal-2020-mad}.
When fine-tuning \xmod{} on a downstream task, the language adapters may be frozen in order to facilitate cross-lingual transfer.
\citet{pfeiffer-etal-2022-lifting} have shown that their approach better preserves monolingual performance.
They have also demonstrated that additional language adapters can be trained after the initial pre-training.

\paragraph{Multilingual Adaptive Pre-training}
The latter can be seen as an instance of adaptive pre-training, i.e., continuing masked language modeling on a corpus of interest.
\citet{alabi-etal-2022-adapting} have shown that such adaptation may be performed simultaneously in many languages.
In addition to adaptation to new languages, downstream tasks can benefit from adaptive pre-training on specific language varieties~\cite{han-eisenstein-2019-unsupervised} or domains~\cite{gururangan-etal-2020-dont}.
Domain adaptation may be performed with data in multiple languages in order to maintain or improve the multilinguality of the model~\cite{kaer-jorgensen-etal-2021-mdapt-multilingual}.

\section{Pre-training Approach}
To create a model that is specialized on the Swiss national languages, we build on a massively multilingual \xmod{} model.\footnote{\url{https://huggingface.co/facebook/xmod-base}}
This model has been pre-trained by \citet{pfeiffer-etal-2022-lifting} on filtered web text in 81 languages, including German, French and Italian.
Our approach combines three ideas from previous work:
\begin{itemize}
    \item \textbf{Domain adaptation:} We continue training the existing language adapters on a large amount of Swiss news articles.
    \item \textbf{Language adaptation:} We train an adapter for the Romansh language.
    \item \textbf{Multilinguality:} We promote transfer between the four languages by using a joint vocabulary and shared embeddings.
\end{itemize}

\begin{figure*}
  \centering
  \includegraphics[width=\textwidth]{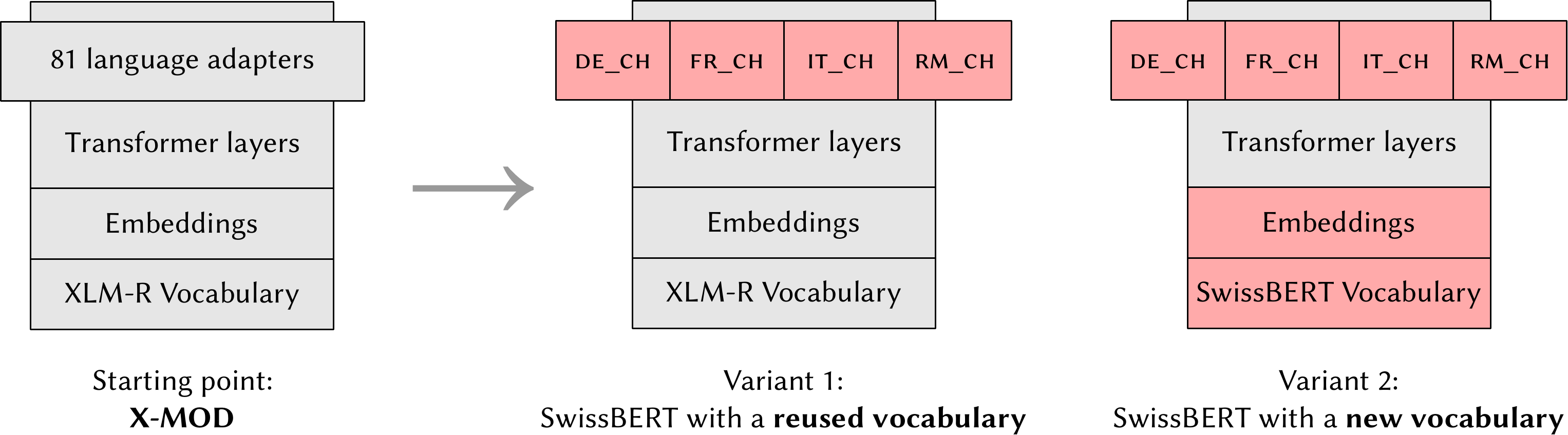}
  \caption{We train two variants of \swissbert{}:
    Variant~1 reuses the vocabulary and embeddings of the pre-trained model, and only language adapters are trained.
    Variant~2 uses a custom \swissbert{} vocabulary based on our pre-training corpus, and multilingual embeddings are trained in addition to the adapters.
  }
    \label{fig:model-variants}
\end{figure*}

\subsection{Pre-training Corpus}\label{subsec:pre-training-corpus}
Our pre-training corpus is composed of media items that have appeared until the end of 2022 and are collected in the Swissdox@LiRI database\footnote{\url{https://swissdox.linguistik.uzh.ch/}}.
The large majority of the items are news articles published in print or in online news portals.
A small part of the items are related types of documents, such as letters to the editor or transcripts of TV news broadcasts.

We retrieve the items directly from the database, which distinguishes our corpus from web-crawled corpora such as the CC100 dataset~\cite{conneau-etal-2020-unsupervised}, on which \xlmr{} and \xmod{} have been trained.
Another difference to CC100 is that our corpus extends to 2022, while the former has been created in or before 2019.
Previous work shows that adaptation to more recent data can improve performance on present-time downstream tasks~\cite{NEURIPS2021_f5bf0ba0}.

We rely on the metadata provided by Swissdox@LiRI to select the articles in the respective languages.
For each language, we hold out articles of the most recent days in the dataset (at least 200 articles) as a validation set.
Like previous work~\cite{NEURIPS2019_c04c19c2,conneau-etal-2020-unsupervised}, we use exponential smoothing to upsample languages with fewer documents, setting $\alpha=0.3$.

\subsection{Modularity}
We follow recommendations by \citet{pfeiffer-etal-2022-lifting} for ensuring the modularity of \swissbert{}.
When pre-training our language adapters, we freeze the shared parameters of the transformer layers.
Conversely, when fine-tuning on downstream tasks, we freeze the language adapters and train the shared parameters.
\citet{pfeiffer-etal-2022-lifting} freeze the embedding layer as well, in order to demonstrate transfer learning across languages with different subword vocabularies.
In this paper, we do not perform experiments of this kind and do not freeze the embedding layer.

\subsection{Vocabulary}
The \xmod{} model reuses the vocabulary of \xlmr{}, which has 250k tokens and has been created based on text in 100 languages~\cite{conneau-etal-2020-unsupervised}.
This presents an interesting trade-off.
On the one hand, \xmod{} already has useful pre-trained multilingual embeddings.
On the other hand, creating a new vocabulary could allow us to represent Switzerland-related words with a smaller degree of segmentation.
This is especially relevant for the Romansh language, which did not contribute to the \xlmr{} vocabulary and as a consequence, is split into many subwords by \xlmr{}:

\textit{Co din ins quai per rumantsch?} \\
\indent \texttt{Co din in|s qua|i per rum|ants|ch ?}

\medskip{}
\noindent To further explore this trade-off, we train two variants of \swissbert{} (Figure~\ref{fig:model-variants}):

\paragraph{Variant 1: reused vocabulary}
We reuse the \xlmr{} vocabulary of \xmod{} and freeze the pre-trained embeddings.
As a consequence, we only train the language adapters.
The other parameters remain identical to \xmod{}.

\paragraph{Variant 2: new vocabulary}
We create a new multilingual vocabulary based on our pre-training corpus.
We follow the procedure of \xlmr{}~\cite{conneau-etal-2020-unsupervised} but restrict the vocabulary size to 50k words.
Specifically, we use SentencePiece~\cite{kudo-richardson-2018-sentencepiece} to create a cased unigram language model~\cite{kudo-2018-subword} with default settings, again smoothing the languages with $\alpha=0.3$.
We then train a new embedding matrix, including new positional embeddings.
Following the recommendation by \citet{pfeiffer-etal-2022-lifting}, we initialize subwords that occur in the original vocabulary with the original embeddings.

\medskip{}
\noindent{}Analyzing the new vocabulary, we find that 18k of the 50k subwords occur in the original \xlmr{} vocabulary, and the other 32k are new subwords.
Appendix~\ref{sec:vocabulary-analysis} lists the new subwords that occur most frequently in the corpus.
Most are Romansh words, orthographic variants, media titles, toponyms or political entities of Switzerland.

\subsection{Preprocessing}
We preprocess the news articles by removing any markup and separating the layout elements, such as headlines, crossheadings, image captions and sidebars, with the special token \texttt{</s>}.
We also remove bylines with author names, photographer names etc., wherever they are marked up as such.

Since previous work has shown that metadata can benefit language modeling~\cite{dhingra-etal-2022-time}, we prefix the articles with their medium and date, for example:

\noindent \footnotesize{\texttt{<medium> rtr.ch <year> 2019 <month> July </s> ...}} \\
\normalsize
where \texttt{<medium>}, \texttt{<year>} and \texttt{<month>} are special tokens.
When training Variant 1, we use the separator symbol instead of custom special tokens:

\noindent \footnotesize{\texttt{</s> rtr.ch </s> 2019 </s> July </s> ...}}

\normalsize

\subsection{Data Analysis}
Additional analysis of the pre-training corpus is provided in the appendices.
Appendix~\ref{sec:discussion-of-data-overlap} shows that there is no relevant overlap with the datasets we use for downstream evaluation.
Appendix~\ref{sec:pre-training-data-statistics} breaks down the number of tokens for each pre-training language, news medium and year of publication.

\subsection{Pre-training Setup}
We generally use the same pre-training setup, implemented in Fairseq~\cite{ott-etal-2019-fairseq}, as was used for \xmod{}.
We make some changes to optimize the efficiency of our pre-training.
Namely, we do not split the articles into sentences but instead train on random contiguous spans of 512 tokens.
In addition, we use a peak learning rate of $7\mathrm{e}{-4}$ throughout.
We train with an effective batch size of 768 across 8 RTX 2080 Ti GPUs.
Both variants of \swissbert{} were trained for 10 epochs.

\begin{table}[!htb]
\begin{tabularx}{\columnwidth}{@{}Xr@{}}
\toprule
Initialization strategy & Validation ppl. \\ \midrule
Italian (\textsc{it\_it})                            & \underline{2.53}\phantom{$\pm$0.0}                  \\
Random initialization               & 2.95$\pm$.13           \\ \bottomrule
\end{tabularx}
\caption{Preliminary experiments for choosing the best initialization of the Italian (\textsc{it\_ch}) language adapter.
We report the standard deviation across three random initializations.
}
\label{tab:italian-initialization-results}
\end{table}

\begin{table}[!htb]
\begin{tabularx}{\columnwidth}{@{}Xr@{}}
\toprule
Initialization strategy & Validation ppl. \\ \midrule
Italian (\textsc{it\_it})                             & 1.85\phantom{$\pm$0.0}                  \\
French (\textsc{fr\_xx})                              & 1.85\phantom{$\pm$0.0}                  \\
German   (\textsc{de\_de})                             & 1.87\phantom{$\pm$0.0}                  \\
\mbox{Average of all Romance languages} & 1.90\phantom{$\pm$0.0}                  \\
Random initialization               & \underline{1.82}$\pm$.02           \\ \bottomrule
\end{tabularx}
\caption{Preliminary experiments for choosing the best initialization of the Romansh language adapter.
The overall perplexity is lower than in Table~\ref{tab:italian-initialization-results} due to the high degree of segmentation when segmenting Romansh text with the \xlmr{} vocabulary.
}
\label{tab:romansh-initialization-results}
\end{table}

\subsection{Initialization of Language Adapters}
In order to choose a strategy for initializing the language adapters, we perform some preliminary experiments based on Variant~1.
Our goal is to train adapters for four language varieties: \textsc{de\_ch}, \textsc{fr\_ch}, \textsc{it\_ch} and \textsc{rm\_ch}.
Three languages already have adapters in \xmod{} -- \textsc{de\_de}, \textsc{fr\_xx} and \textsc{it\_it} -- and so we expect that the best result can be achieved by continuing training these adapters.

We verify this hypothesis on the example of Italian.
Table~\ref{tab:italian-initialization-results} shows the validation perplexity of the model after pre-training on the Italian part of our corpus for 2k steps.
An adapter initialized with \xmod{}'s Italian adapter yields a lower perplexity than a randomly initialized adapter.
Thus, domain-adaptive (and variety-adaptive) pre-training seems more efficient than training an adapter from scratch.

In the case of Romansh, we similarly hypothesize that initializing from Italian or another Romance language will outperform a randomly initialized adapter, given the relatedness of these languages.
However, Table~\ref{tab:romansh-initialization-results} shows that random initialization yields a lower perplexity for Romansh.
In addition, averaging multiple language adapters -- e.g., the adapters for all the Romance languages in \xmod{}
-- is clearly not a viable strategy.
\pagebreak
Given these findings, we opt for the following initialization strategy:
\begin{itemize}
    \item \textsc{de\_ch} from \textsc{de\_de};
    \item \textsc{fr\_ch} from \textsc{fr\_xx};
    \item \textsc{it\_ch} from \textsc{it\_it};
    \item \textsc{rm\_ch} from scratch.
\end{itemize}

\section{Evaluation}
For evaluating \swissbert{}, we focus on Switzerland-related natural language understanding tasks, and especially multilingual and cross-lingual tasks on the token or sequence level.

\subsection{Tasks}

\paragraph{Named Entity Recognition (NER)}
Our main question is whether \swissbert{} has improved natural language understanding capabilities in the domain it has been adapted to.
To evaluate this, we annotate named entities in contemporary news articles and test whether a \swissbert{} model fine-tuned on NER can detect the entities with higher accuracy than baseline models.

We name our test set \swissner{}.\footnote{\url{https://huggingface.co/datasets/ZurichNLP/SwissNER}}
Specifically, we annotate 200 paragraphs per language that we extracted from publicly accessible articles by the Swiss Broadcasting Corporation~(SRG SSR).
The annotated articles have been published in February 2023 and are thus not contained in the pre-training corpus.
Appendices~\ref{sec:swissner-annotation-process} and \ref{sec:swissner-data-statistics} describe the dataset in detail.

For fine-tuning on the NER task we use \wikineural{}, an automatically labeled dataset in nine languages~\cite{tedeschi-etal-2021-wikineural-combined}.
Only the data in German, French and Italian are relevant to \swissbert{}, and so we train the model jointly on these three parts of \wikineural{}.
As a consequence, when training baselines on \wikineural{}, we report separate results for training only on German, French and Italian, and for training on all the nine languages.

Since \wikineural{} does not contain training data in Romansh, we evaluate zero-shot transfer to this language.
In the case of \xmod{}, we activate the Italian adapter when performing inference on Romansh.

\paragraph{NER on Historical News}
In addition to contemporary news, we report results for two datasets from the HIPE-2022 shared task~\cite{ehrmann2022overview}.
Other than \swissner{}, this task involves NER on mostly historical, OCR-processed news articles:

\begin{itemize}
    \item \texttt{hipe2020}: We fine-tune and evaluate on annotated Swiss and Luxembourgish newspaper articles from the \textit{Impresso} collection~\cite{ehrmann-etal-2020-language} that are written in French or German, ranging between the years 1798 and 2018.
    \item \texttt{letemps}: We fine-tune and evaluate on annotated newspaper articles from two Swiss newspapers in French~\cite{ehrmann2016diachronic}, ranging between 1804 and 1981.
\end{itemize}

\begin{table*}[!htb]
\begin{adjustbox}{max width=\textwidth}
\begin{tabular}{@{}lrrrr@{}} \toprule
                             & Supervised \textsc{de\_ch} & Supervised \textsc{fr\_ch} & Supervised \textsc{it\_ch}  & Zero-shot \textsc{rm\_ch}      \\ \midrule
\mbox{\xlmr{}~\cite{conneau-etal-2020-unsupervised}}  & & & &  \\
\mbox{– fine-tuned on 9 languages}  & 70.7$\pm$1.0 & 70.9$\pm$0.6 & 76.6$\pm$1.2 & 63.8$\pm$0.7 \\
\mbox{– fine-tuned on \textsc{de}, \textsc{fr}, \textsc{it}}  & 71.7$\pm$0.7 & 70.5$\pm$0.2 & 76.7$\pm$0.7 & 64.6$\pm$0.7 \\
\mbox{\xmod{}~\cite{pfeiffer-etal-2022-lifting}}  &  & & &  \\
\mbox{– fine-tuned on 9 languages}  & 71.2$\pm$0.7 & 70.4$\pm$0.3 & 75.9$\pm$0.9 &  61.5$\pm$0.7 \\
\mbox{– fine-tuned on \textsc{de}, \textsc{fr}, \textsc{it}}  & 72.2$\pm$0.5 & 71.8$\pm$1.1 & 76.7$\pm$0.8 & 61.4$\pm$1.8 \\ \midrule
\mbox{\swissbert{} (fine-tuned on \textsc{de}, \textsc{fr}, \textsc{it})}  &  &  &  &  \\
\mbox{– reused vocabulary}  & 74.5$\pm$0.8 & 74.2$\pm$0.9 & 78.6$\pm$0.1 & 81.8$\pm$0.9 \\
\mbox{– new vocabulary}     & \underline{74.8}$\pm$1.2 & \underline{75.9}$\pm$0.8 & \underline{79.2}$\pm$0.5 & \underline{83.7}$\pm$0.9 \\ \bottomrule
\end{tabular}
\end{adjustbox}
\caption{Named entity recognition results on the \swissner{} test set.
The last column reports zero-shot results for Romansh.
Since \xmod{} does not have a Romansh adapter, we use the Italian adapter when applying \xmod{} to the Romansh test set.
The best results are underlined.
}
\label{tab:swissner-results}
\end{table*}

\begin{table*}[!htb]
\begin{adjustbox}{max width=\textwidth}
\begin{tabular}{@{}lrrrrrr@{}} \toprule
                             & Coarse & Coarse & Coarse & Fine & Fine & Fine \\
                             & \footnotesize{\texttt{hipe2020} \textsc{fr}}            & \footnotesize{\texttt{hipe2020} \textsc{de}}           & \footnotesize{\texttt{letemps} \textsc{fr}}  & \footnotesize{\texttt{hipe2020} \textsc{fr}}            & \footnotesize{\texttt{hipe2020} \textsc{de}}           & \footnotesize{\texttt{letemps} \textsc{fr}}      \\ \midrule
French Europeana BERT~\cite{stefan_schweter_2020_4275044}  & \underline{81.2}$\pm$0.4 & - & \underline{68.3}$\pm$1.7   & \underline{75.9}$\pm$0.6 & - & \underline{63.0}$\pm$1.2 \\
German Europeana BERT~\cite{stefan_schweter_2020_4275044}  & - & \underline{76.1}$\pm$0.7 & -   & - & \underline{68.2}$\pm$1.0 & - \\
\xlmr{}~\cite{conneau-etal-2020-unsupervised}                         & 79.3$\pm$1.1 & 72.7$\pm$1.5 & 66.1$\pm$1.2 & 73.6$\pm$1.3 & 64.4$\pm$0.8 & 60.6$\pm$1.0  \\
\xmod{}~\cite{pfeiffer-etal-2022-lifting}                         & 77.2$\pm$1.1 & 69.0$\pm$2.1 & 63.5$\pm$1.1 & 70.2$\pm$1.1 & 58.9$\pm$2.4 & 58.1$\pm$1.1  \\ \midrule
\swissbert{}                       &  &  &    &  &  &   \\
\mbox{– reused vocabulary}  & 77.7$\pm$1.3 & 69.2$\pm$1.9 & 64.3$\pm$1.1 & 71.7$\pm$1.1 & 58.8$\pm$1.1 & 57.6$\pm$1.1  \\
\mbox{– new vocabulary}     & 80.0$\pm$1.4 & 71.6$\pm$1.9 & 66.2$\pm$1.1 & 73.4$\pm$1.0 & 62.2$\pm$1.7 & 60.4$\pm$1.4  \\ \bottomrule
\end{tabular}
\end{adjustbox}
\caption{Named entity recognition on historical newspapers~(HIPE-2022,~\cite{ehrmann2022overview}).
We report a strict micro-averaged F1-score for the coarse tag set (left) and the fine-grained tag set (right).
}
\label{tab:hipe-results}
\end{table*}

\paragraph{Stance Detection}
Another source of domain shift, apart from historical text, could be user-generated text.
We evaluate our models on multilingual stance detection with the \xstance{} dataset~\cite{vamvas-sennrich-2020-xstance}, which is based on comments written by Swiss political candidates.
The dataset contains 67k comments on various political issues in either German, French or Italian.
Given a question and a comment, the task is to judge whether the candidate has taken a stance in favor or against the issue at hand.
We follow \citet{vamvas-sennrich-2020-xstance} and use the concatenation of the two sequences as an input to \swissbert{}:

\texttt{<s> [question] </s></s> [comment] </s>}

\noindent The model is then trained to predict a binary label for the sequence pair based on the hidden state for \texttt{<s>}.

\paragraph{Sentence Retrieval}
To further investigate \swissbert{}'s ability to align text in the Romansh language to the other languages, we construct a sentence retrieval task out of a German--Romansh parallel corpus of 597 unique sentence pairs~\cite{dolev-2023-romansh}.
This task is inspired by parallel corpus mining tasks~\cite{zweigenbaum-etal-2017-overview} and the Tatoeba test set used by \citet{artetxe-schwenk-2019-massively}.

Specifically, we use the German sentences as queries and report top-1 accuracy when retrieving the corresponding Romansh sentences.
As similarity metric we use \bertscore{}~\cite{zhang2020bertscore}, which allows us to use the pre-trained models directly without any fine-tuning.\footnote{Note that calculating \bertscore{} for all pairs of sentences is viable in the context of this experiment but would not be efficient for large-scale parallel corpus mining.}
While \citet{zhang2020bertscore} recommend using a validation set to determine the best transformer layer for \bertscore{}, we opt for a simpler approach and use the average hidden states across all layers.

The German--Romansh sentence pairs have been sampled by \citet{dolev-2023-romansh} from press releases published by the Canton of the Grisons between 1997 and 2022.\footnote{\url{https://github.com/eyldlv/DERMIT-Corpus}}
Most of the releases were originally written in German and then manually translated into Romansh Grischun~\cite{scherrer-cartoni-2012-trilingual}.
The gold sentence alignment is based on an automatic alignment that has been manually verified by a trained linguist.

\paragraph{Word Alignment}
Finally, we evaluate \swissbert{} on German--Romansh word alignment using the unsupervised \mbox{SimAlign} technique~\cite{jalili-sabet-etal-2020-simalign}.
For testing we use the same parallel sentences as above, which have been manually annotated with word alignments by a trained linguist~\cite{dolev-2023-romansh}.
We predict word alignments using the ``Match'' variant of SimAlign and report the F1-score with regard to the gold annotations.
We do not perform a grid search to find the optimal layer but instead average the hidden states across all transformer layers.

\subsection{Baseline Models}
\paragraph{General-purpose models}
\begin{itemize}
    \item \xlmr{}, a model trained jointly on 100 languages~\cite{conneau-etal-2020-unsupervised}
    \item \xmod{}, a model trained with language adapters on 81 languages, which is the basis of \swissbert{}~\cite{pfeiffer-etal-2022-lifting}
\end{itemize}

\paragraph{Specialized models}
\begin{itemize}
    \item Europeana BERT models pre-trained on historical newspapers in the German or French language~\cite{stefan_schweter_2020_4275044}
\end{itemize}

\begin{table*}[!htb]
\begin{adjustbox}{max width=\textwidth}
\begin{tabular}{@{}lrrrrr@{}} \toprule
                             & Supervised \textsc{de} & Supervised \textsc{fr} & Cross-topic \textsc{de} & Cross-topic \textsc{fr} & Cross-lingual \textsc{it} \\ \midrule
\xlmr{}~\cite{conneau-etal-2020-unsupervised} & 76.9$\pm$0.8 &	78.6$\pm$0.8 &	73.0$\pm$1.3 &	75.3$\pm$1.9 &	74.4$\pm$0.7               \\
\xmod{}~\cite{pfeiffer-etal-2022-lifting} & 77.5$\pm$0.7 &	78.5$\pm$0.7 &	73.6$\pm$0.7 &	74.5$\pm$0.8 &	74.7$\pm$0.8                  \\ \midrule
\swissbert{}                       &    &    &                &                &                  \\
\mbox{– reused vocabulary}                & 77.9$\pm$0.4 &	79.2$\pm$0.3 &	73.8$\pm$0.5 &	74.5$\pm$0.8 &	74.8$\pm$0.6 \\
\mbox{– new vocabulary}                    & \underline{78.3}$\pm$0.4 &	\underline{80.1}$\pm$0.5 &	\underline{74.0}$\pm$0.6 &	\underline{75.8}$\pm$0.5 &	\underline{74.9}$\pm$0.7 \\ \bottomrule
\end{tabular}
\end{adjustbox}
\caption{Stance detection on political comments in the \xstance{} dataset~\cite{vamvas-sennrich-2020-xstance}.
We report the F1-score for different test sets of \xstance{}.
}
\label{tab:xstance-results}
\end{table*}

\begin{table}[]
\begin{adjustbox}{max width=\columnwidth}
\begin{tabular}{@{}lrr@{}}
\toprule
         &  Sentence retrieval & Word alignment  \\ \midrule
\xlmr{}~\cite{conneau-etal-2020-unsupervised} & 25.3 &   62.6             \\
\xmod{}~\cite{pfeiffer-etal-2022-lifting} &  31.8   &   65.1            \\ \midrule
\swissbert{}   &       &            \\
\mbox{– reused vocabulary}    & 92.0   & 85.9              \\
\mbox{– new vocabulary}   & \underline{95.6}      & \underline{86.4}           \\ \bottomrule
\end{tabular}
\end{adjustbox}
\caption{German–Romansh parallel corpus alignment: sentence retrieval accuracy and word alignment F1-score across 597 sentence pairs.
}
\label{tab:retrieval-results}
\end{table}

\pagebreak

\subsection{Fine-tuning}
We try to avoid hyperparameter optimization and instead use settings from previous work that are known to work well for \xlmr{} and similar models.
\begin{itemize}
    \item For fine-tuning on \wikineural{}, we train the models for 3 epochs with a learning rate of $2\mathrm{e}{-5}$ and a batch size of 16.\footnote{\url{https://huggingface.co/Babelscape/wikineural-multilingual-ner}}
          We report the average and standard deviation across 5 random seeds.
    \item For fine-tuning on the HIPE-2022 datasets, we use a learning rate of $5\mathrm{e}{-5}$ and a batch size of 8~\cite{ehrmann2022overview}.
          However, we train for up to 25 epochs to ensure that all models converge.
          We report the average and standard deviation across 10 random seeds.
    \item For fine-tuning on \xstance{}, we train with a learning rate of $1\mathrm{e}{-5}$ and a batch size of 16 for 3 epochs, with a maximum sequence length of 256 tokens~\cite{schick-schutze-2021-exploiting}.
          We report the average and standard deviation across 10 random seeds.
\end{itemize}
\indent We implement fine-tuning with the \textit{transformers} library~\cite{wolf-etal-2020-transformers} and otherwise use the default settings of the library.

\subsection{Results}
Evaluation results are presented in Tables~\ref{tab:swissner-results}–\ref{tab:retrieval-results}.
Overall, we find that \swissbert{} outperforms the baselines on Switzerland-related tasks, and especially on Romansh.

Furthermore, the results show that using a custom vocabulary when adapting \xmod{} is beneficial, not only for Romansh but also for the three languages that are represented in the original \xlmr{} vocabulary.
One reason could be that the custom vocabulary better matches the evaluation domain.
Another reason could be that the model has more capacity to adapt to the target domain if the embedding layer is trained in addition to the language adapters, irrespective of the vocabulary.

An interesting comparison is NER on contemporary news~
(Table~\ref{tab:swissner-results}) and historical news~(Table~\ref{tab:hipe-results}).
While \swissbert{} outperforms the baselines on contemporary news, the model is not consistently better than \xlmr{} on historical news.
On the latter task, \swissbert{} strongly improves over the non-adapted model, \xmod{}, but inherits the low baseline performance of \xmod{} compared to \xlmr{}.
One explanation why \xlmr{} outperforms \xmod{} on historical NER is that it was trained for more steps with a larger batch size.
Secondly, \xlmr{} does not depend on language identification, which might be beneficial when training on historical or OCR-processed text.
We find that monolingual models trained on historical news surpass general-purpose multilingual models, confirming previous findings~\cite{ehrmann2022overview,ryser-et-al-2022-exploring}.

The \xstance{} task is informative because it is based on user-generated text, as opposed to newspaper articles.
\swissbert{} moderately but systematically outperforms the baselines on this task (Table~\ref{tab:xstance-results}), which indicates that it could be a useful model for processing not only news, but Switzerland-related text in general.

Finally, the German--Romansh alignment experiment~(Table~\ref{tab:retrieval-results}) demonstrates that self-supervised training is sufficient to enable multilingual representations for Romansh Grischun, despite the rather small pre-training corpus.
\swissbert{} strongly outperforms multilingual encoders that have not been specifically trained on Romansh.
We expect that \swissbert{} could be a valuable resource for future Romansh NLP applications, such as classification, retrieval, or parallel corpus alignment.

\section{Conclusion}
We release a language model that supports the four national languages of Switzerland.
Specific challenges of the Swiss language situation are addressed using methods from the recent literature, including multilingual masked language modeling, language adapters, and adaptive pre-training.
We evaluate the resulting model, which we call \swissbert{}, on a range of Switzerland-related natural language understanding tasks and mostly see an improved accuracy.
In addition, \swissbert{} excels in tasks involving Romansh, compared to models that do not cover this language.

\subsection*{Limitations}
The \swissbert{} model and our evaluation experiments have a limited scope.
First of all, the training objective of \swissbert{} limits the range of direct applications.
\swissbert{} is mainly intended for tagging tokens in written text (e.g., named entity recognition, part-of-speech tagging), text classification, and the encoding of words, sentences or documents into fixed-size embeddings.
\swissbert{} is not designed for generating text.

Secondly, we expect \swissbert{} to perform best on input that is similar to our pre-training corpus of written news.
Switzerland also has language varieties that are rarely found in newspapers, e.g., Swiss German and dialects of Romansh.
While these are currently not covered by \swissbert{}, the model is designed to be extensible.

Finally, the main goal of our evaluation experiments is to verify that the adaptation of \swissbert{} has been effective, i.e., that \swissbert{} has a higher accuracy on Switzerland-related tasks than non-adapted baselines.
We do not methodically compare different approaches.
In this paper, we present one approach that we have found to work well, but further ablation experiments would be required to verify that it is the optimal approach.

\section*{Acknowledgements}
This work was funded by the Swiss National Science Foundation (project MUTAMUR; no.~176727).
It makes use of media data made available via Swissdox@LiRI by the Linguistic Research Infrastructure of the University of Zurich (see \url{https://t.uzh.ch/1hI} for more information).
We thank Pedro Ortiz Suarez for early feedback and Eyal Dolev, Maud Ehrmann, Sven Najem-Meyer and Jonas Pfeiffer for help with downstream evaluation.

\bibliography{bibliography}

\appendix

\bigskip
\bigskip

\section{Model Card}


\subsection{Model Details}

\subsubsection{Model Description}


\begin{itemize}
  \item Model type: \xmod{}~\cite{pfeiffer-etal-2022-lifting}.
  \item Languages: German, French, Italian, Romansh.
  \item License: Attribution-NonCommercial 4.0 International (CC BY-NC 4.0).
  \item Fine-tuned from model: \texttt{xmod-base}
\end{itemize}

\subsubsection{Model Sources}


\begin{itemize}
  \item Source code:\\ \url{https://github.com/ZurichNLP/swissbert}
  \item Model weights:\\ \url{https://huggingface.co/ZurichNLP/swissbert}
  \item Backup:\\ \url{https://doi.org/10.5281/zenodo.8016844}
\end{itemize}

%

%
%
%
%
%

\subsection{Bias, Risks, and Limitations}


%
\begin{itemize}
    \item The model was adapted on written news articles and might perform worse on other domains or language varieties.
    \item While we have removed many author bylines, we did not anonymize the pre-training corpus. The model might have memorized information that has been described in the news but is no longer in the public interest.
\end{itemize}

\subsection{Training Details}

\subsubsection{Training Data}

German, French, Italian and Romansh documents in the Swissdox@LiRI database, until 2022~(Section~\ref{subsec:pre-training-corpus}).

\subsubsection{Training Procedure}
Masked language modeling~\cite{devlin-etal-2019-bert, conneau-etal-2020-unsupervised}.


%
%
%

\subsection{Environmental Impact}


\begin{itemize}
  \item Hardware type: RTX 2080 Ti
  \item Hours used: 2 models $\times$ 10 epochs $\times$ 18 hours $\times$ 8 devices = 2880 hours
  \item Site: Zurich, Switzerland
  \item Energy source: 100\% hydropower\footnote{\label{note1}Source: \url{https://t.uzh.ch/1rU}}
  \item Carbon efficiency: 0.0016 kg CO\textsubscript{2}e/kWh\textsuperscript{\ref{note1}}
  \item Carbon emitted: 1.15 kg CO\textsubscript{2}e~\cite{lacoste2019quantifying}
\end{itemize}

%
%

\vfill

\pagebreak

\section{\swissner{} Annotation Process}\label{sec:swissner-annotation-process}
\swissner{} is a dataset for named entity recognition based on manually annotated news articles in Swiss Standard German, French, Italian, and Romansh Grischun.
We annotate a selection of articles that have been published in February 2023 on the following online news portals:
\begin{itemize}
    \item German: \url{https://www.srf.ch/}
    \item French: \url{https://www.rts.ch/}
    \item Italian: \url{https://www.rsi.ch/}
    \item Romansh: \url{https://www.rtr.ch/}
\end{itemize}
The four portals belong to the Swiss Broadcasting Corporation~(SRG SSR).
We select news articles in the categories ``Switzerland'' or ``Regional''.
The articles in the individual languages are not translations of each other and tend to cover different regions of Switzerland, but the editing style and the overall topics are coherent.

For each article we extract the first two paragraphs after the lead paragraph.
We follow the guidelines of the CoNLL-2002 and 2003 shared tasks~\cite{tjong-kim-sang-2002-introduction,tjong-kim-sang-de-meulder-2003-introduction} and annotate the names of persons, organizations, locations and miscellaneous entities.
The annotation was performed by a single annotator.

\vfill

\pagebreak

\section{Discussion of Data Overlap}\label{sec:discussion-of-data-overlap}
Below we analyze the data overlap between pre-training and downstream evaluation:
\begin{itemize}
    \item \swissner{} dataset: None of the articles are in the pre-training corpus, which does not contain articles from 2023.
    \item \texttt{hipe2020}: No overlap.
    \item \texttt{letemps}: No overlap.
    \item \xstance{}: The dataset does not contain news.
    \item German–Romansh parallel corpus:
    \begin{itemize}
        \item German: 36 out of 597 sentences appear verbatim in the pre-training corpus.
        \item Romansh: 23 out of 597 sentences appear verbatim in the pre-training corpus.
    \end{itemize}
\end{itemize}
Note that the German sentences and Romansh sentences never appear together in the pre-training corpus, making it unlikely that overlap gives \swissbert{} an advantage in the alignment task.

We also make sure to exclude articles that occur in the CHeeSE dataset~\cite{mascarell-etal-2021-stance} to facilitate future evaluation on this dataset.

\vfill
\pagebreak
\onecolumn

\section{Model Sizes}\label{sec:model-sizes}
\begin{table*}[htb!]
\begin{adjustbox}{max width=\textwidth}
\begin{tabular}{@{}lrrrr@{}}
\toprule
                                        & Adapters & Vocabulary & Parameters & Trained parameters (adaptation) \\ \midrule
\xlmr{}~\cite{conneau-etal-2020-unsupervised} & - & 250\,002 & 278\,043\,648 & - \\
\xmod{}~\cite{pfeiffer-etal-2022-lifting} & 81 & 250\,002 & 852\,472\,320 & - \\ \midrule
\swissbert{} &  &  &  & \\
 – reused vocabulary & 4                  & 250\,002          & 306\,410\,496        & 28\,366\,848                   \\
 – new vocabulary    & 4                  & 50\,262           & 153\,010\,176        & 67\,163\,136                   \\ \bottomrule
\end{tabular}
\end{adjustbox}
\caption{Sizes of the models used in the experiments. The second variant of \swissbert{} has fewer parameters due to the smaller vocabulary, but has more trained parameters because we train the embedding layer.}
\label{tab:model-sizes}
\end{table*}

\vfill

\section{\swissner{} Data Statistics}\label{sec:swissner-data-statistics}
\begin{table}[htb!]
\begin{tabularx}{\textwidth}{@{}Xrrrrr@{}}
\toprule
                     & \textsc{de\_ch} & \textsc{fr\_ch} & \textsc{it\_ch} & \textsc{rm\_ch} & Total \\ \midrule
Number of paragraphs & 200 & 200 & 200 & 200 & 800 \\
Number of tokens & 9\,498 & 11\,434 & 12\,423 & 13\,356 & 46\,711 \\
Number of entities & 479 & 475 & 556 & 591 & 2\,101 \\
– PER & 104 & 92 & 93 & 118 & 407 \\
– ORG & 193 & 216 & 266 & 227 & 902 \\
– LOC & 182 & 167 & 197 & 246 & 792 \\
– MISC & 113 & 79 & 88 & 39 & 319 \\ \bottomrule
\end{tabularx}
\caption{Statistics for the \swissner{} test sets.}
\label{tab:swissner-statistics}
\end{table}

\vfill

\section{Additional Baselines for \swissner{}}
\begin{table*}[htb!]
\begin{adjustbox}{max width=\textwidth}
\begin{tabular}{@{}lrrrr@{}} \toprule
                             & Supervised \textsc{de\_ch} & Supervised \textsc{fr\_ch} & Supervised \textsc{it\_ch}  & Zero-shot \textsc{rm\_ch}      \\ \midrule
\texttt{wikineural-multilingual-ner}~\cite{tedeschi-etal-2021-wikineural-combined}  & 71.4\phantom{$\pm$0.0} & 71.4\phantom{$\pm$0.0} & 75.2\phantom{$\pm$0.0} & 66.5\phantom{$\pm$0.0}  \\
German Europeana BERT~\cite{stefan_schweter_2020_4275044}  & 67.6$\pm$0.5 & 64.4$\pm$1.2 & 68.6$\pm$1.1 & 58.8$\pm$0.6  \\
French Europeana BERT~\cite{stefan_schweter_2020_4275044}  & 57.0$\pm$1.2 & 69.0$\pm$0.7 & 66.8$\pm$0.7 & 61.0$\pm$1.1  \\ \midrule
\mbox{\swissbert{}} (new vocabulary)  & 74.8$\pm$1.2 & 75.9$\pm$0.8 & 79.2$\pm$0.5 & 83.7$\pm$0.9  \\ \bottomrule
\end{tabular}
\end{adjustbox}
\caption{Results for additional baselines on the \swissner{} test set.
\texttt{wikineural-multilingual-ner} is an mBERT model fine-tuned by \citet{tedeschi-etal-2021-wikineural-combined} on the \wikineural{} dataset.
The other models in the table have been fine-tuned on the German, French and Italian parts of \wikineural{}.
}
\label{tab:additional-swissner-results}
\end{table*}

\vfill
\vfill
\vfill
\vfill

\clearpage

\section{Pre-training Data Statistics}\label{sec:pre-training-data-statistics}

\bigskip

\begin{table*}[htb!]
\begin{adjustbox}{max width=\textwidth}
\begin{tabular}{@{}lrrrrr@{}}
\toprule
                 & \textsc{de\_ch} & \textsc{fr\_ch} & \textsc{it\_ch} & \textsc{rm\_ch} & Total \\ \midrule
\textit{Training set}  &      &  &    &   &     \\
Number of articles  & 17\,832\,421     & 3\,681\,679     & 48\,238     & 32\,750     & 21\,595\,088     \\
Number of tokens: &    &    &    &    &       \\
– in terms of \xlmr{} vocabulary & 11\,611\,859\,339 & 2\,651\,272\,875 & 27\,504\,679 & 16\,977\,167 & 14\,307\,614\,060 \\
– in terms of the new \swissbert{} vocabulary    & 9\,857\,117\,034  & 2\,384\,955\,915 & 26\,825\,471 & 13\,286\,172 & 12\,282\,184\,592 \\ \midrule
\textit{Validation set}  &      &  &    &   &     \\
Number of articles  & 1\,401     & 263     & 214    & 211     & 2\,089     \\
Number of tokens: &    &    &    &    &       \\
– in terms of \xlmr{} vocabulary & 1\,648\,604 & 256\,794 & 95\,088 & 348\,166 & 2\,348\,652 \\
– in terms of the new \swissbert{} vocabulary    & 1\,416\,928 & 234\,498 & 93\,450 & 267\,904 & 2\,012\,780 \\ \bottomrule
\end{tabular}
\end{adjustbox}
\caption{Number of articles and number of subword tokens in our pre-training data.}
\label{tab:token-statistics}
\end{table*}

\bigskip
\bigskip
\bigskip

\begin{figure*}[htb!]
  \centering
  \includegraphics[width=\textwidth]{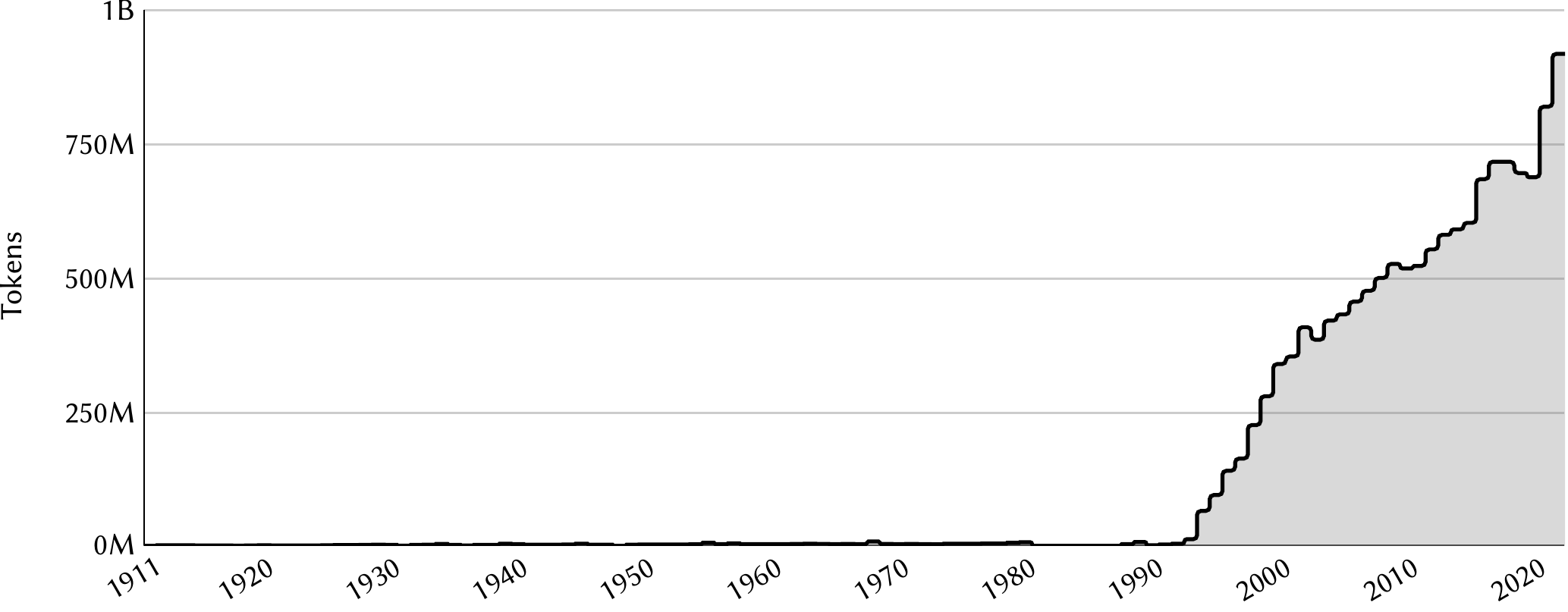}
  \caption{Number of tokens (in terms of \xlmr{} vocabulary) per year in the training set.
  }
    \label{fig:years-statistics}
\end{figure*}

\begin{table*}[htb!]
\footnotesize
\begin{adjustbox}{height=0.7\textwidth}
\begin{tabular}{@{}lrrc@{}}
\toprule
Medium                             & Articles & Tokens & Lang. \\ \midrule
Neue Zürcher Zeitung               & 1\,189\,914          & 891\,991\,094        & \textsc{de}                \\
St. Galler Tagblatt                & 1\,297\,896          & 660\,251\,490        & \textsc{de}                \\
Tages-Anzeiger                     & 932\,263            & 598\,824\,374        & \textsc{de}                \\
Berner Zeitung                     & 999\,254            & 539\,180\,275        & \textsc{de}                \\
Neue Luzerner Zeitung              & 950\,663            & 494\,536\,537        & \textsc{de}                \\
Der Bund                           & 720\,637            & 492\,383\,793        & \textsc{de}                \\
nzz.ch                             & 495\,217            & 472\,250\,494        & \textsc{de}                \\
Aargauer Zeitung / MLZ             & 697\,225            & 422\,696\,456        & \textsc{de}                \\
Basler Zeitung                     & 784\,388            & 408\,674\,904        & \textsc{de}                \\
Le Temps                           & 400\,971            & 398\,369\,886        & \textsc{fr}                \\
Tribune de Genève                  & 508\,101            & 344\,341\,310        & \textsc{fr}                \\
cash.ch                            & 536\,877            & 312\,495\,330        & \textsc{de}                \\
Blick                              & 584\,893            & 236\,515\,780        & \textsc{de}                \\
Schweizer Illustrierte             & 239\,447            & 230\,676\,599        & \textsc{de}                \\
tagesanzeiger.ch                   & 249\,453            & 226\,827\,823        & \textsc{de}                \\
Zürichsee-Zeitung                  & 453\,383            & 224\,321\,469        & \textsc{de}                \\
tdg.ch                             & 273\,969            & 208\,280\,959        & \textsc{fr}                \\
srf.ch                             & 307\,414            & 203\,538\,182        & \textsc{de}                \\
Le Matin                           & 316\,890            & 195\,390\,209        & \textsc{fr}                \\
Der Landbote                       & 371\,054            & 195\,316\,528        & \textsc{de}                \\
bernerzeitung.ch                   & 224\,923            & 179\,355\,942        & \textsc{de}                \\
24 heures                          & 251\,709            & 166\,651\,833        & \textsc{fr}                \\
20 minuten online                  & 262\,474            & 165\,019\,108        & \textsc{de}                \\
20 minutes online                  & 242\,166            & 159\,347\,598        & \textsc{fr}                \\
Thurgauer Zeitung                  & 328\,194            & 154\,509\,265        & \textsc{de}                \\
SonntagsZeitung                    & 181\,129            & 153\,405\,631        & \textsc{de}                \\
24 Heures                          & 190\,122            & 151\,106\,511        & \textsc{fr}                \\
Finanz und Wirtschaft              & 174\,529            & 149\,271\,153        & \textsc{de}                \\
24heures.ch                        & 179\,011            & 146\,908\,146        & \textsc{fr}                \\
Soloth. Zeitung / MLZ              & 242\,015            & 144\,028\,122        & \textsc{de}                \\
lematin.ch                         & 222\,071            & 139\,517\,644        & \textsc{fr}                \\
blick.ch                           & 234\,082            & 138\,524\,249        & \textsc{de}                \\
Oltner Tagblatt / MLZ              & 209\,641            & 132\,205\,560        & \textsc{de}                \\
derbund.ch                         & 140\,174            & 129\,723\,367        & \textsc{de}                \\
Die Weltwoche                      & 92\,430             & 129\,170\,717        & \textsc{de}                \\
Zofinger Tagblatt / MLZ            & 236\,835            & 127\,415\,159        & \textsc{de}                \\
Blick.ch                           & 242\,403            & 126\,481\,730        & \textsc{de}                \\
NZZ am Sonntag                     & 149\,052            & 120\,989\,713        & \textsc{de}                \\
tagblatt.ch                        & 59\,245             & 111\,287\,915        & \textsc{de}                \\
Sonntagsblick                      & 163\,860            & 106\,270\,536        & \textsc{de}                \\
srf Video                          & 15\,981             & 105\,429\,243        & \textsc{de}                \\
bazonline.ch                       & 105\,805            & 98\,151\,941         & \textsc{de}                \\
Le Matin Dimanche                  & 98\,321             & 88\,226\,207         & \textsc{fr}                \\
Solothurner Zeitung                & 170\,381            & 87\,820\,588         & \textsc{de}                \\
HandelsZeitung                     & 83\,784             & 81\,119\,111         & \textsc{de}                \\
\mbox{Basellandsch. Zeit. / MLZ} & 129\,362            & 78\,326\,422         & \textsc{de}                \\
Aargauer Zeitung                   & 156\,674            & 75\,069\,605         & \textsc{de}                \\
fuw.ch                             & 84\,644             & 73\,232\,927         & \textsc{de}                \\
20 minuten                         & 297\,464            & 70\,811\,524         & \textsc{de}                \\
Luzerner Zeitung                   & 107\,555            & 68\,626\,196         & \textsc{de}                \\
Zürcher Unterländer                & 126\,689            & 68\,400\,789         & \textsc{de}                \\
L'Hebdo                            & 57\,388             & 67\,428\,263         & \textsc{fr}                \\
Newsnetz                           & 108\,330            & 66\,200\,248         & \textsc{de}                \\
Cash                               & 67\,878             & 64\,433\,800         & \textsc{de}                \\
Die Wochenzeitung                  & 49\,112             & 60\,875\,629         & \textsc{de}                \\
\mbox{Werdenberger \& Obertogg.}   & 99\,981             & 56\,790\,265         & \textsc{de}                \\
24 heures Région La Côte           & 90\,197             & 56\,212\,161         & \textsc{fr}                \\
20 minutes                         & 240\,986            & 55\,868\,424         & \textsc{fr}                \\
L'Illustré                         & 53\,216             & 55\,536\,046         & \textsc{fr}                \\
letemps.ch                         & 37\,911             & 50\,789\,072         & \textsc{fr}                \\
Blick am Abend                     & 152\,891            & 49\,614\,784         & \textsc{de}                \\
Schweizer Familie                  & 56\,387             & 49\,099\,500         & \textsc{de}                \\
Glückspost                         & 83\,685             & 48\,476\,534         & \textsc{de}                \\
Mittelland Zeitung                 & 91\,380             & 46\,214\,129         & \textsc{de}                \\ \bottomrule
\end{tabular}
\end{adjustbox}
\begin{adjustbox}{height=0.7\textwidth}
\begin{tabular}{@{}lrrc@{}}
\toprule
Medium                             & Articles & Tokens & Lang. \\ \midrule
Facts                              & 31\,044             & 45\,951\,513         & \textsc{de}                \\
Limmattaler Zeit. / MLZ            & 73\,310             & 44\,858\,755         & \textsc{de}                \\
luzernerzeitung.ch                 & 49\,806             & 44\,347\,311         & \textsc{de}                \\
Berner Rundschau / MLZ             & 78\,420             & 43\,671\,698         & \textsc{de}                \\
aargauerzeitung.ch                 & 31\,223             & 43\,405\,747         & \textsc{de}                \\
RTS.ch                             & 76\,210             & 39\,938\,988         & \textsc{fr}                \\
handelszeitung.ch                  & 52\,064             & 39\,558\,669         & \textsc{de}                \\
Handelszeitung                     & 34\,270             & 36\,153\,858         & \textsc{de}                \\
\mbox{Zentralschweiz am Sonntag}   & 46\,753             & 35\,613\,607         & \textsc{de}                \\
Zuger Zeitung                      & 52\,369             & 35\,471\,227         & \textsc{de}                \\
\mbox{Schweiz am Sonntag / MLZ}    & 45\,374             & 33\,968\,154         & \textsc{de}                \\
AZ-Tabloid / MLZ                   & 79\,120             & 33\,694\,480         & \textsc{de}                \\
landbote.ch                        & 31\,026             & 33\,333\,601         & \textsc{de}                \\
rts.ch                             & 41\,202             & 33\,213\,658         & \textsc{fr}                \\
\mbox{Limmattaler Tagblatt / MLZ}         & 65\,911             & 33\,025\,066         & \textsc{de}                \\
Das Magazin                        & 17\,676             & 32\,874\,311         & \textsc{de}                \\
rts Vidéo                          & 3\,460              & 31\,243\,548         & \textsc{fr}                \\
Schweiz am Wochenende              & 40\,152             & 30\,710\,690         & \textsc{de}                \\
Beobachter                         & 33\,005             & 30\,659\,313         & \textsc{de}                \\
Urner Zeitung                      & 40\,906             & 29\,990\,931         & \textsc{de}                \\
Bilanz                             & 25\,174             & 29\,777\,904         & \textsc{de}                \\
Tele                               & 40\,704             & 29\,728\,436         & \textsc{de}                \\
Sonntag / MLZ                      & 41\,560             & 29\,016\,500         & \textsc{de}                \\
langenthalertagblatt.ch            & 23\,436             & 28\,866\,148         & \textsc{de}                \\
www.sf.tv                          & 60\,662             & 28\,385\,473         & \textsc{de}                \\
zsz.ch                             & 25\,174             & 28\,250\,302         & \textsc{de}                \\
zuonline.ch                        & 24\,308             & 27\,908\,168         & \textsc{de}                \\
berneroberlaender.ch               & 23\,401             & 27\,692\,709         & \textsc{de}                \\
thunertagblatt.ch                  & 23\,525             & 27\,621\,041         & \textsc{de}                \\
Le Nouveau Quotidien               & 31\,310             & 27\,192\,253         & \textsc{fr}                \\
Berner Oberländer                  & 41\,331             & 27\,179\,816         & \textsc{de}                \\
Wiler Zeitung                      & 43\,265             & 26\,514\,799         & \textsc{de}                \\
Appenzeller Zeitung                & 42\,908             & 26\,419\,047         & \textsc{de}                \\
Toggenburger Tagblatt              & 41\,470             & 25\,560\,534         & \textsc{de}                \\
20 Minuten                         & 123\,750            & 25\,065\,828         & \textsc{de}                \\
bzbasel.ch                         & 19\,618             & 24\,377\,477         & \textsc{de}                \\
\mbox{bz - Zeit. f.d. Region Basel}  & 32\,265             & 24\,078\,982         & \textsc{de}                \\
Obwaldner Zeitung                  & 31\,427             & 23\,589\,402         & \textsc{de}                \\
Nidwaldner Zeitung                 & 31\,273             & 23\,002\,223         & \textsc{de}                \\
schweizer‐illustrierte.ch          & 29\,438             & 22\,230\,522         & \textsc{de}                \\
SWI swissinfo.ch                   & 16\,978             & 21\,915\,894         & \textsc{de}                \\
Bilan                              & 13\,993             & 20\,086\,032         & \textsc{fr}                \\
züritipp (Tages-Anzeiger)          & 39\,076             & 19\,358\,626         & \textsc{de}                \\
20 Minutes                         & 90\,470             & 18\,526\,178         & \textsc{fr}                \\
Badener Tagblatt                   & 24\,518             & 18\,210\,357         & \textsc{de}                \\
TV 8                               & 35\,231             & 16\,991\,076         & \textsc{fr}                \\
Ostschweiz am Sonntag              & 25\,472             & 16\,521\,054         & \textsc{de}                \\
badenertagblatt.ch                 & 18\,079             & 16\,382\,179         & \textsc{de}                \\
Der Sonntag / MLZ                  & 20\,582             & 15\,563\,949         & \textsc{de}                \\
swissinfo.ch                       & 9\,484              & 15\,276\,786         & \textsc{de}                \\
www.swissinfo.ch                   & 11\,373             & 14\,884\,339         & \textsc{fr}                \\
\textbf{rtr.ch}                             & \textbf{32\,622}             & \textbf{14\,746\,036}         & \textbf{\textsc{rm}}                \\
solothurnerzeitung.ch              & 15\,475             & 14\,063\,626         & \textsc{de}                \\
Thuner Tagblatt                    & 19\,570             & 13\,719\,511         & \textsc{de}                \\
\textbf{rsi.ch}                             & \textbf{36\,741}             & \textbf{12\,871\,992}         & \textbf{\textsc{it}}                \\
dimanche.ch                        & 16\,770             & 12\,617\,761         & \textsc{fr}                \\
PME Magazine                       & 10\,371             & 12\,481\,323         & \textsc{fr}                \\
\mbox{BZ - Langenthaler Tagblatt}         & 15\,538             & 12\,326\,791         & \textsc{de}                \\
Grenchner Tagblatt / MLZ           & 22\,587             & 12\,194\,558         & \textsc{de}                \\
\mbox{24 h. Région Nord Vaudois}      & 23\,573             & 12\,012\,095         & \textsc{fr}                \\
Grenchner Tagblatt                 & 16\,022             & 11\,822\,579         & \textsc{de}                \\
\mbox{24 h. Régon Riviera Chablais}   & 22\,803             & 11\,703\,359         & \textsc{fr}                \\
24 h. Région Lausannoise       & 20\,208             & 10\,093\,139         & \textsc{fr}                \\
grenchnertagblatt.ch               & 11\,506             & 10\,062\,085         & \textsc{de}                \\ \bottomrule
\end{tabular}
\end{adjustbox}
\caption{Number of articles and number of tokens (in terms of \xlmr{} vocabulary) per news medium in the training set. Note that some media statistics are distributed over multiple variants of the title. We report the majority language for each medium and highlight the two media that have majority language Italian and Romansh, respectively. The table does not include media with fewer than 10M tokens in the training corpus.}
\label{tab:media-statistics}
\end{table*}

\vfill
\clearpage

\section{Vocabulary Analysis}\label{sec:vocabulary-analysis}
\begin{table}[!htbp]
\footnotesize
\begin{tabularx}{0.35\textwidth}{@{}lXr@{}}
\toprule
Rank & Subword            & Majority language \\ \midrule
\textsc{126} & \texttt{.»}              & \textsc{de} \\
\textsc{163} & \texttt{\_Franken}        & \textsc{de} \\
\textsc{302} & \texttt{\_francs}         & \textsc{fr} \\
\textsc{335} & \texttt{\_betg}           & \textsc{rm} \\
\textsc{359} & \texttt{\_ins}            & \textsc{de} \\
\textsc{387} & \texttt{\_Zürcher}        & \textsc{de} \\
\textsc{403} & \texttt{\_rsi}            & \textsc{it} \\
\textsc{405} & \texttt{\_rtr}            & \textsc{rm} \\
\textsc{428} & \texttt{\_Berner}         & \textsc{de} \\
\textsc{497}  & \texttt{\_Tagblatt}       & \textsc{de} \\
\textsc{516}  & \texttt{\_quai}           & \textsc{rm} \\
\textsc{545}  & \texttt{MLZ}             & \textsc{de} \\
\textsc{589}  & \texttt{\_èn}             & \textsc{rm} \\
\textsc{628}  & \texttt{\_Galler}         & \textsc{de} \\
\textsc{639}  & \texttt{\_Lausanne}       & \textsc{fr} \\
\textsc{654}  & \texttt{\_Gemeinderat}    & \textsc{de} \\
\textsc{698}  & \texttt{\_Luzerner}       & \textsc{de} \\
\textsc{699}  & \texttt{\_grossen}        & \textsc{de} \\
\textsc{701}  & \texttt{strasse}         & \textsc{de} \\
\textsc{706}  & \texttt{\_heisst}         & \textsc{de} \\
\textsc{710}  & \texttt{\_Basler}         & \textsc{de} \\
\textsc{732}  & \texttt{\_onns}           & \textsc{rm} \\
\textsc{741}  & \texttt{\_SVP}            & \textsc{de} \\
\textsc{748}  & \texttt{\_suisse}         & \textsc{fr} \\
\textsc{778}  & \texttt{\_Tribune}        & \textsc{fr} \\
\textsc{779}  & \texttt{\_anc}            & \textsc{rm} \\
\textsc{785}  & \texttt{\_Svizra}         & \textsc{rm} \\
\textsc{807}  & \texttt{\_Matin}          & \textsc{fr} \\
\textsc{816}  & \texttt{\_persunas}       & \textsc{rm} \\
\textsc{824}  & \texttt{\_Quai}           & \textsc{rm} \\
\textsc{840}  & \texttt{\_dentant}        & \textsc{rm} \\
\textsc{918}  & \texttt{\_vegn}           & \textsc{rm} \\
\textsc{924}  & \texttt{\_Aargauer}       & \textsc{de} \\
\textsc{964}  & \texttt{\_Lugano}         & \textsc{it} \\
\textsc{971}  & \texttt{\_Bundesrat}      & \textsc{de} \\
\textsc{991}  & \texttt{Anzeiger}        & \textsc{de} \\
\textsc{1011}  & \texttt{onn}             & \textsc{rm} \\
\textsc{1026}  & \texttt{\_Grischun}       & \textsc{rm} \\
\textsc{1033}  & \texttt{\_Luzern}         & \textsc{de} \\
\textsc{1049}  & \texttt{sda}             & \textsc{de} \\
\textsc{1098}  & \texttt{\_RTR}            & \textsc{rm} \\
\textsc{1102}  & \texttt{\_canton}         & \textsc{fr} \\
\textsc{1106}  & \texttt{\_könne}          & \textsc{de} \\
\textsc{1112}  & \texttt{\_Sieg}           & \textsc{de} \\
\textsc{1146}  & \texttt{Nous}            & \textsc{fr} \\
\textsc{1163}  & \texttt{\_Gemeinden}      & \textsc{de} \\
\textsc{1186}  & \texttt{\_Temps}          & \textsc{fr} \\
\textsc{1199}  & \texttt{\_erklärte}       & \textsc{de} \\
\textsc{1208}  & \texttt{\_FDP}            & \textsc{de} \\
\textsc{1218}  & \texttt{\_Dass}           & \textsc{de} \\ \bottomrule
\end{tabularx}
\hspace*{0.05\textwidth}
\begin{tabularx}{0.35\textwidth}{@{}lXr@{}}
\toprule
Rank & Subword            & Majority language \\ \midrule
\textsc{1239} & \texttt{\_Fussball}       & \textsc{de} \\
\textsc{1244} & \texttt{\_Spital}         & \textsc{de} \\
\textsc{1268} & \texttt{\_vegnir}         & \textsc{rm} \\
\textsc{1271} & \texttt{\_cunter}         & \textsc{rm} \\
\textsc{1274} & \texttt{\_Covid}          & \textsc{fr} \\
\textsc{1279} & \texttt{\_Keystone}       & \textsc{de} \\
\textsc{1280} & \texttt{\_Gallen}         & \textsc{de} \\
\textsc{1300} & \texttt{\_Aktien}         & \textsc{de} \\
\textsc{1303} & \texttt{\_Cussegl}        & \textsc{rm} \\
\textsc{1323} & \texttt{\_liess}          & \textsc{de} \\
\textsc{1327} & \texttt{\_WM}             & \textsc{de} \\
\textsc{1335} & \texttt{\_uschia}         & \textsc{rm} \\
\textsc{1340} & \texttt{\_kommenden}      & \textsc{de} \\
\textsc{1363} & \texttt{\_Kantons}        & \textsc{de} \\
\textsc{1380} & \texttt{\_Federer}        & \textsc{de} \\
\textsc{1426} & \texttt{\_chantun}        & \textsc{rm} \\
\textsc{1434} & \texttt{\_sajan}          & \textsc{rm} \\
\textsc{1438} & \texttt{\_erstmals}       & \textsc{de} \\
\textsc{1439} & \texttt{\_Thun}           & \textsc{de} \\
\textsc{1442} & \texttt{\_dapli}          & \textsc{rm} \\
\textsc{1445} & \texttt{\_Stimmen}        & \textsc{de} \\
\textsc{1452} & \texttt{\_tranter}        & \textsc{rm} \\
\textsc{1454} & \texttt{\_dix}            & \textsc{fr} \\
\textsc{1459} & \texttt{\_fatg}           & \textsc{rm} \\
\textsc{1465} & \texttt{\_suenter}        & \textsc{rm} \\
\textsc{1491} & \texttt{\_milliards}      & \textsc{fr} \\
\textsc{1496} & \texttt{\_Strassen}       & \textsc{de} \\
\textsc{1520} & \texttt{\_Mitteilung}     & \textsc{de} \\
\textsc{1543} & \texttt{\_Entscheid}      & \textsc{de} \\
\textsc{1555} & \texttt{\_Urs}            & \textsc{de} \\
\textsc{1559} & \texttt{\_Massnahmen}     & \textsc{de} \\
\textsc{1563} & \texttt{\_Zurich}         & \textsc{fr} \\
\textsc{1576} & \texttt{\_müsse}          & \textsc{de} \\
\textsc{1577} & \texttt{\_Behörden}       & \textsc{de} \\
\textsc{1580} & \texttt{\_Stadtrat}       & \textsc{de} \\
\textsc{1590} & \texttt{\_zeigte}         & \textsc{de} \\
\textsc{1594} & \texttt{\_Regierungsrat}  & \textsc{de} \\
\textsc{1603} & \texttt{\_Kantonspolizei} & \textsc{de} \\
\textsc{1604} & \texttt{\_machte}         & \textsc{de} \\
\textsc{1615} & \texttt{\_Mrd}            & \textsc{de} \\
\textsc{1644} & \texttt{\_gemäss}         & \textsc{de} \\
\textsc{1646} & \texttt{\_schliesslich}   & \textsc{de} \\
\textsc{1648} & \texttt{\_Thurgauer}      & \textsc{de} \\
\textsc{1659} & \texttt{\_Amt}            & \textsc{de} \\
\textsc{1686} & \texttt{\_tdg}            & \textsc{fr} \\
\textsc{1690} & \texttt{\_Solothurner}    & \textsc{de} \\
\textsc{1697} & \texttt{\_duai}           & \textsc{rm} \\
\textsc{1705} & \texttt{\_UBS}            & \textsc{de} \\
\textsc{1716} & \texttt{\_Cun}            & \textsc{rm} \\
\textsc{1719}  & \texttt{\_CVP}            & \textsc{de} \\ \bottomrule
\end{tabularx}
\caption{The 100 most frequent subwords that appear in our custom \swissbert{} vocabulary but not in the \xlmr{} vocabulary.
The symbol \texttt{\_} (U+2581) is used by SentencePiece to denote preceding whitespace.
For each subword we report the majority language, i.e., the language that contributes the subword more often than the other languages, after exponential smoothing.
Out of the 100 top subwords, 26 originate from Romansh and usually have a functional meaning, e.g., `not', `one', `this', or `in'.
Other subwords can be explained by Swiss orthographic conventions, such as the use of `ss' in place of `ß' in Swiss Standard German or the use of outward-pointing guillemets without surrounding whitespace (`.»').
Most remaining subwords are~(parts of) media titles, toponyms or political entities.
Given that \xlmr{} was created in or before 2019, the neologism \texttt{\_Covid} belongs to the words that only occur in the \swissbert{} vocabulary.
}
\label{tab:top-subwords}
\vspace{-50pt}
\end{table}

\end{document}